\documentclass{article}

\usepackage{arxiv}

\usepackage[utf8]{inputenc} % allow utf-8 input
\usepackage[T1]{fontenc}    % use 8-bit T1 fonts
\usepackage{hyperref}       % hyperlinks
\usepackage{url}            % simple URL typesetting
\usepackage{booktabs}       % professional-quality tables
\usepackage{amsfonts}       % blackboard math symbols
\usepackage{nicefrac}       % compact symbols for 1/2, etc.
\usepackage{microtype}      % microtypography
\usepackage{lipsum}		% Can be removed after putting your text content
\usepackage{graphicx}
\usepackage{natbib}
\usepackage{doi}

\usepackage{graphicx}
\usepackage{amsmath}
\usepackage{multirow}
\usepackage{float}
\usepackage{array}

\title{Anonymizing medical case-based explanations through disentanglement}

\author{ Helena Montenegro\thanks{Corresponding author: \texttt{maria.h.sampaio@inesctec.pt}},\hspace{1mm} Jaime S. Cardoso \\
	Faculty of Engineering of the University of Porto, Porto, Portugal \\
	INESC TEC, Porto, Portugal
}

\date{}

\renewcommand{\shorttitle}{Anonymizing medical case-based explanations through disentanglement}

\hypersetup{
pdftitle={A template for the arxiv style},
pdfsubject={q-bio.NC, q-bio.QM},
pdfauthor={David S.~Hippocampus, Elias D.~Striatum},
pdfkeywords={First keyword, Second keyword, More},
}

\begin{document}
\maketitle

\begin{abstract}
Case-based explanations are an intuitive method to gain insight into the decision-making process of deep learning models in clinical contexts. However, medical images cannot be shared as explanations due to privacy concerns. To address this problem, we propose a novel method for disentangling identity and medical characteristics of images and apply it to anonymize medical images. The disentanglement mechanism replaces some feature vectors in an image while ensuring that the remaining features are preserved, obtaining independent feature vectors that encode the images' identity and medical characteristics. We also propose a model to manufacture synthetic privacy-preserving identities to replace the original image's identity and achieve anonymization. The models are applied to medical and biometric datasets, demonstrating their capacity to generate realistic-looking anonymized images that preserve their original medical content. Additionally, the experiments show the network's inherent capacity to generate counterfactual images through the replacement of medical features.
\end{abstract}

% keywords can be removed

\section{Introduction}

% Interpretability of Deep Learning models
The applicability of Deep Learning models in health is restricted by the trust that medical experts and patients have in the models. 
For the models to be trusted, they should provide explanations that support their decisions, enabling humans to understand their reasoning. 
Case-based explanations, which show examples from the data as explanations, are particularly intuitive for medical experts since medical image retrieval systems are often included in the clinical practice \citep{silva2020interpretability}. 
Recently, various works have been claiming that it is possible to re-identify patients based on medical images, such as brain Magnetic Resonance Images (MRIs) \citep{esmeral2022low}, and chest radiographs \citep{packhauser2022deep}. As such, medical images cannot be shared as explanations without compromising the privacy of patients, requiring anonymization.

The anonymization of medical case-based explanations must consider the characteristics of medical data. Private medical datasets from health institutions may contain a small number of images per patient due to the difficulty of the image acquisition process associated with obtaining radiographs or MRIs. Furthermore, the anonymization process must preserve the intelligibility of the images, so that humans can understand them, and the medical content of the images, to not compromise their value as explanations \citep{montenegro2022privacy}. Most works in the image anonymization literature are not applied to medical images and do not consider the explicit preservation of explanatory evidence \citep{montenegro2021icml,chen2018vganbased,cho2020cleanir,gong2020disentangled,oleszkiewicz2019siamese}. The anonymization methods specifically developed for medical case-based explanations present limitations in the preservation of the medical content of the images and in the realism of the anonymized images \citep{montenegro2021privacy, montenegro2023disentangled}. 

% Summary of proposed method
To address the limitations in the literature, we propose a deep generative model capable of disentangling biometric and medical characteristics from medical images. The model is an autoencoder that decomposes images into three independent vectors that capture identity features, medical features, and the remaining features required to reconstruct the original image. The mechanism to ensure that the extracted feature vectors are independent relies on the definition of independence: two variables are independent if an alteration to one variable does not affect the other one. We apply this reasoning to build a loss function that ensures that an alteration to one of the feature vectors of the original image does not cause alterations to the remaining vectors in the generated image. To ensure that the images generated by the disentanglement network are realistic even with limited training data, we use an adversarial loss typical of Generative Adversarial Networks (GANs) \citep{goodfellow2014gan} in addition to a reconstruction loss, and apply augmentations to the generated data before providing it to the model's discriminator \citep{karras2020training}. After separating the image into independent vectors of identity and medical features, the network can easily be applied to image anonymization and to the generation of counterfactual explanations, by replacing the identity and medical features of the original image, respectively. We use a Variational Autoencoder (VAE) \citep{kingma2014vae} to generate vectors of synthetic identities to replace the identities in the images and achieve anonymization.

The proposed network is applied to three datasets. We use a biometric dataset of eye iris images (Warsaw-BioBase-Disease-Iris v2.1 from \citet{trokielewicz2015iris,trokielewicz2017iris}), to verify the network's capacity to anonymize images with well-defined identity features. We use a medical dataset of chest radiographs (CheXpert from  \citet{irvin2019chexpert}) to verify the model's capacity to disentangle and preserve medical features. Moreover, we apply the network to a dataset of face images (CelebAMask-HQ from \citet{lee2020maskgan}), to verify the network's disentanglement capacity in images that contain well-defined identities and attributes, enabling a facilitated visualization of the network's strengths.

The contributions of this work are the following:
\begin{enumerate}
    \item The proposal of a novel approach to disentangle identity and medical characteristics from medical images, capable of modifying identity features while preserving its medical features. The model is also inherently capable of generating counterfactual explanations through the replacement of medical features.
    \item The proposal of a model to generate privacy-preserving synthetic identities that can be used to replace the identities of the original image using the proposed disentanglement model, obtaining anonymized images whose medical content is preserved.
    \item The extensive evaluation of the proposed models, including a comparison between the models and the state-of-the-art medical image anonymization methods, and ablation studies that verify how the performance of the disentanglement network is affected by alterations to its loss function and training.
\end{enumerate}

The remaining sections are structured as follows: Section~\ref{sec:related-work} presents an overview of the related work on image anonymization. Section~\ref{sec:approach} describes in detail the proposed disentanglement and anonymization models. Section~\ref{sec:experiments} describes the experimental setup, including datasets, training details and the evaluation methodology. Section~\ref{sec:results} presents and discusses the results. Finally, Section~\ref{sec:conclusions} concludes the paper.

\section{Related Work}  \label{sec:related-work}

This section analyses the related work in image anonymization techniques, considering their application to medical case-based explanations.
Traditional image anonymization methods such as blurring, pixelation, and the K-Same-based methods \citep{newton2005ksame,gross2006ksame}, which average several images from different identities, fail to preserve realism and explanatory evidence \citep{montenegro2021icml}.

Deep Learning methods use deep generative networks to generate realistic anonymized images that follow the probability distribution of real data. The most common generative models used in the image anonymization literature are Generative Adversarial Networks (GANs) \citep{goodfellow2014gan} and Variational Autoencoders (VAEs) \citep{kingma2014vae}. GANs model the probability distribution of the data implicitly through a generator network that is trained to generate images such that an adversarial discriminator that distinguishes between real and synthetic images confuses them with real images. VAEs model the probability distribution by transforming it into a simpler distribution using an encoder-decoder architecture. Most methods in the literature use identity recognition networks to guide the training of the generative model to obtain anonymized images. Nonetheless, there are some exceptions, like DeepPrivacy \citep{hukkelaas2019deepprivacy, hukkelaas2023deepprivacy2}, which do not use identity recognition and do not guarantee that the generated image's identity is sufficiently different from the training data to not compromise data privacy.

Some existing deep image anonymization methods use disentangled representation learning techniques to explicitly obtain identity features disentangled from the remaining image features. These methods replace the identity features of the original image with a synthetic vector of features.
The $R^2VAE$ model \citep{gong2020disentangled} disentangles identity-related and identity-independent features in face images using a cycle-consistency loss that replaces the identity in an image and then restores it. On inference, the identity is replaced by synthetic features obtained by averaging identity features from several training images. The CLEANIR model \citep{cho2020cleanir} uses a VAE to separate face images into identity-related and identity-independent features. However, this method does not promote independence between the two feature vectors, failing to disentangle them. The synthetic features are obtained by applying the Gram-Schmidt process to the identity features of the original image. Neither of these works considers preserving features related to a specific task, failing to preserve explanatory evidence if applied to case-based explanations.

Other methods alter an image's identity without explicitly obtaining it, typically through a loss function. PPRL-VGAN \citep{chen2018vganbased} replaces the identity in face images using a GAN that generates images that an identity recognition network recognizes as belonging to the target identity. CIAGAN \citep{maximov2020ciagan} replaces the identity in images by conditioning a GAN into generating images with the identity embeddings of a randomly selected training sample. SGAP \citep{oleszkiewicz2019siamese} and PP-GAN \citep{wu2019privacy} use GANs to maximize the identity-related distance between the original and the anonymized images. Although PPRL-VGAN preserves the class (facial expression) of the original image during anonymization, all these methods fail to preserve the exact task-related features of the original images that would serve as explanatory evidence in the context of explanations.

Only two works have been applied to anonymize medical and biometric case-based explanations, considering the explicit preservation of explanatory evidence.
The GANs proposed by \citet{montenegro2021privacy} anonymize images that an identity recognition network has difficulty classifying, by approximating its predictions to a uniform distribution, and by maximizing the identity-related distance between the anonymized and original images. 
To preserve medical features, it reconstructs the disease-related features of the original images, obtained through saliency maps, in the anonymized images. However, this method preserves the absolute position of the medical features, rather than adapting them to the alterations that occur to the structure of the image as a result of the anonymization process. The disentanglement model proposed by \citet{montenegro2023disentangled} extracts disease-invariant identity features and identity-invariant medical features from images. The adversarial training used to obtain identity-invariant medical features forces the network to capture the minimum amount of medical features required to perform the disease recognition task that do not contain the patient's identity. As such, there may be more medical features that may be discarded during anonymization. In the same manner, there may also be identity features that may be preserved through the anonymization process. Moreover, the images generated by both these works lack the intelligibility required to apply them in real clinical contexts.

In this work, we propose a novel method to disentangle identity and medical features and to anonymize images preserving the medical content and intelligibility of the images. In the experiments, we compare the proposed model with the two works from the literature that have been applied to anonymize case-based explanations, demonstrating the proposed network's superiority at anonymizing images.

Table~\ref{tab:gap-analysis} compares the deep image anonymization methods regarding their capacity to anonymize medical case-based explanations. In this analysis, we consider that methods that use identities from the training data during anonymization (by averaging them and using them as a replacement) do not preserve privacy for all the patients in the data.

\begin{table*}[t]
\caption{Comparison between deep image anonymization methods from the scientific literature.}
\label{tab:gap-analysis}
\centering
\begin{tabular}{|p{0.255\linewidth}|>{\centering\arraybackslash}p{0.09\linewidth}|>{\centering\arraybackslash}p{0.04\linewidth}|>{\centering\arraybackslash}p{0.065\linewidth}|>{\centering\arraybackslash}p{0.15\linewidth}|>{\centering\arraybackslash}p{0.11\linewidth}|>{\centering\arraybackslash}p{0.07\linewidth}|}
\hline
\multirow{2}{*}{Method} & Applied to medical & \multicolumn{2}{>{\centering\arraybackslash}p{0.13\linewidth}|}{Preserves semantic} & Preserves privacy for original & Preserves privacy for & Realistic images \\ \cline{3-4}
& data & class & features & patient & all & \\ \hline
\citep{hukkelaas2019deepprivacy} &  &  &  & $\times$ &  & \\ \hline
\citep{hukkelaas2023deepprivacy2} &  &  &  & $\times$ &  & $\times$ \\ \hline
\citep{gong2020disentangled} &  &  &  & $\times$ &  & $\times$ \\ \hline
\citep{cho2020cleanir} &  &  &  & $\times$ & $\times$ & $\times$ \\ \hline
\citep{maximov2020ciagan} &  &  &  & $\times$ &  & $\times$ \\ \hline
\citep{oleszkiewicz2019siamese} &  &  &  & $\times$ &  & \\ \hline
\citep{wu2019privacy}  &  &  &  & $\times$ &  & $\times$ \\ \hline
\citep{chen2018vganbased} &  & $\times$ &  & $\times$ &  & $\times$ \\ \hline
\citep{montenegro2021privacy} &  & $\times$ & $\times$ & $\times$ & $\times$ & \\ \hline
\citep{montenegro2023disentangled} & $\times$ & $\times$ & $\times$ & $\times$ & $\times$ & \\ \hline
\textbf{Ours} & $\times$ & $\times$ & $\times$ & $\times$ & $\times$ & $\times$ \\ \hline
\end{tabular}
\end{table*}

\section{Proposed Methodology} \label{sec:approach}

The following subsections describe the models proposed for disentangling identity and medical features and synthesizing privacy-preserving identity features for anonymization.

\subsection{Disentanglement}

\begin{figure*}[t]
    \centering
    \includegraphics[scale=0.65]{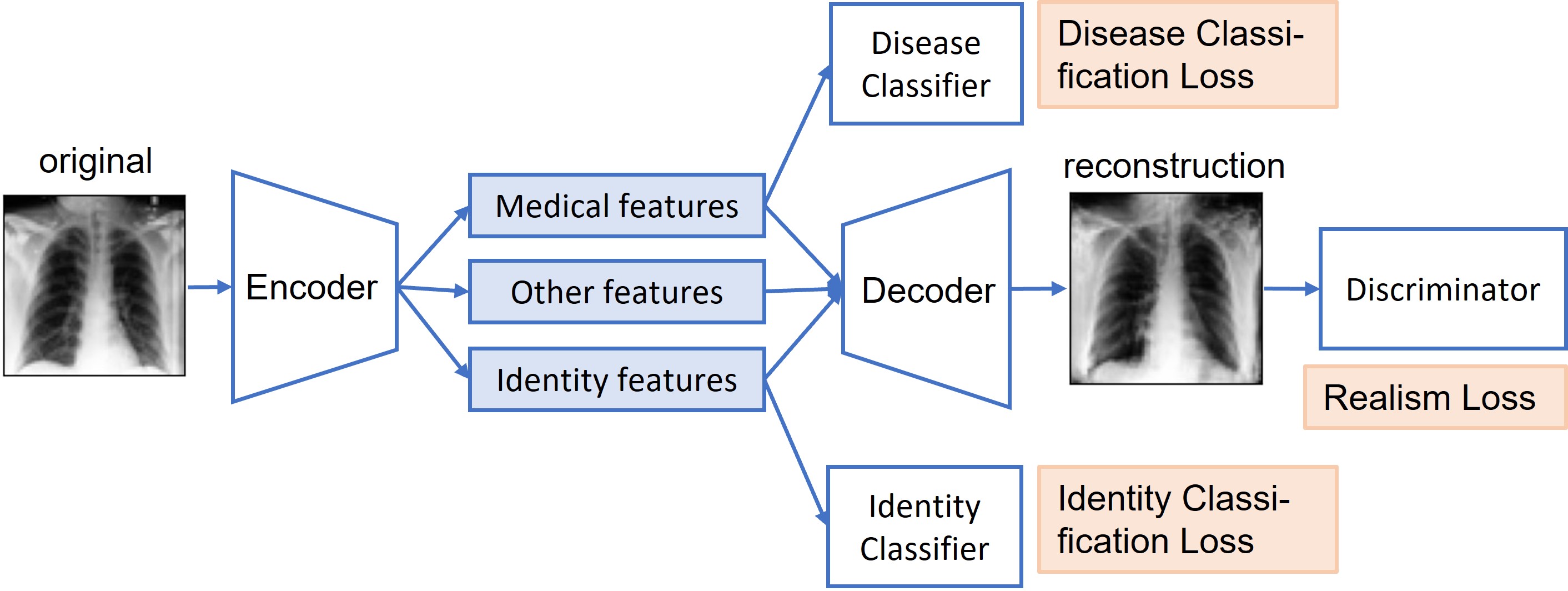}
    \caption{Overview of the generative network without disentanglement loss.} \label{fig:generative-network}
\end{figure*}

We propose a generative model for disentangling identity and medical characteristics in images.
The disentanglement network is an autoencoder that extracts a representation of an image into three independent vectors encoding its identity features, medical features, and remaining features, and then obtains a new image based on this latent representation. An overview of the model is introduced in Figure~\ref{fig:generative-network}. The model is composed of the following networks:

\begin{itemize}
    \item \textbf{Encoder ($E$)}: extracts features from the original image into three vectors: medical features, identity features, and the remaining image features required to reconstruct the original image.
    \item \textbf{Decoder ($D$)}: generates an image based on the feature vectors. It is trained to reconstruct the original image when its feature vectors are provided.
    \item \textbf{Disease Classifier ($C_{med}$)}: classifies the pathology based on the medical features extracted by the encoder. Its purpose is to guarantee that the vector of medical features captures the features of the image required to perform the medical predictive task.
    \item \textbf{Identity Classifier ($C_{id}$)}: identifies the patient in the image based on its identity features. Its purpose on the network is to ensure that the identity feature vector captures the features that reveal the identity of patients.
    \item \textbf{Discriminator ($C_{real}$)}: classifies the images generated by the decoder based on whether they look real or fake. It is the discriminator of a typical GAN whose main purpose is to increase the realism of the images generated by the autoencoder.
\end{itemize}

In our implementation, the encoder, the decoder, and the discriminator are convolutional neural networks with residual connections, following the architecture of a ResNet \citep{he2016deep}. The disease and identity classifiers are multi-layer perceptions with one hidden layer, a dropout layer, and a final classification layer.

The network has three goals: separate an image into three independent feature vectors, ensure that two of the vectors encode medical and identity features, respectively, and obtain realistic synthetic images. The goals of the network are achieved through its three loss terms.

\begin{figure*}[t]
    \centering
    \includegraphics[scale=0.6]{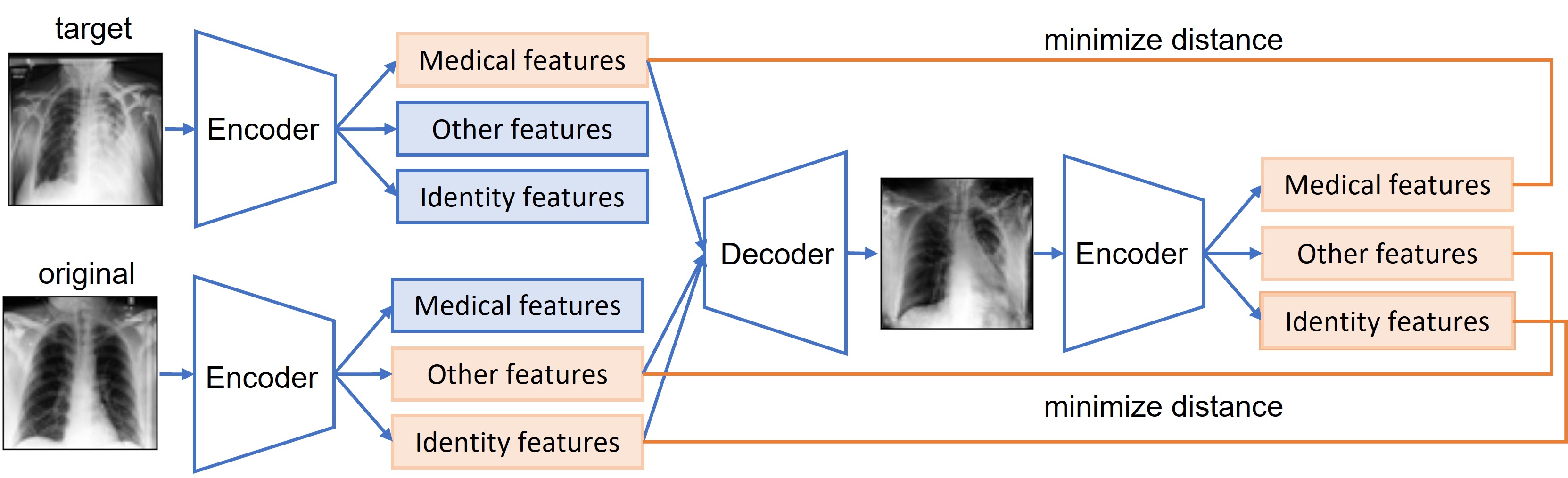}
    \caption{Representation of the disentanglement process through the replacement of medical features of the original image.} \label{fig:disentanglement-process}
\end{figure*}

The disentanglement loss term promotes independence between vectors.
If two or more vectors are independent, then making alterations to one of the vectors should not cause alterations to the remaining vectors. As such, we implement a disentanglement mechanism in the network that replaces one of the vectors of the original image and promotes that no alterations occur in the remaining vectors, by minimizing the distance between the remaining vectors of the original and altered images, as exemplified in Figure~\ref{fig:disentanglement-process}. The target image whose features are used as a replacement must contain a different pathology and identity from the original image. The disentanglement loss is exposed in Equation~\ref{eq:disentanglement-loss}, where $I_{ori}$, $I_{tar}$, and $I_{gen}$ represent the original image, the target image, and the image generated by replacing the feature vector $i$ of the original image by the respective feature vector of the target image, respectively. $E_{i}$ and $E_{j}$ represent the feature vectors extracted by the encoder that are replaced ($i$) and not replaced ($j$) during the disentanglement, respectively.

\begin{eqnarray}
\begin{aligned}
\mathcal{L}_{disent} = E_{I \sim p_d(I)}[ \sum_{i} ((E_i(I_{tar}) - E_i(I_{gen}))^2 + \sum_{j \neq i} (E_j(I_{ori}) - E_j(I_{gen}))^2)] \label{eq:disentanglement-loss}
\end{aligned}
\end{eqnarray}

The classification loss ensures that two of the independent feature vectors extracted by the encoder capture medical and identity features, respectively.
A disease recognition network is used to capture the medical features of the image in one of the feature vectors. We use an identity classifier to ensure the encoder extracts identity features into one of its vectors.
The classification loss is shown in Equation~\ref{eq:multiclass-classifier}, where $I$ represents the original image, $E_{med}$ and $E_{id}$ represent the medical and identity features extracted by the encoder, respectively, $y_{med}$ and $y_{id}$ represent the ground-truth classification annotations, in the form of one-hot encodings, for the medical and identity recognition tasks, respectively, and $\lambda_{med}$ and $\lambda_{id}$ are parameters to control the importance of each loss term.

\begin{eqnarray}
\begin{aligned}
\mathcal{L}_{class} = E_{I \sim p_d(I)}[-\lambda_{med} y_{med}\log(C_{med}(E_{med}(I))) - \lambda_{id} y_{id}\log(C_{id}(E_{id}(I)))] \label{eq:multiclass-classifier}
\end{aligned}
\end{eqnarray}

The realism loss promotes realism in the images generated by the decoder. The decoder is trained to reconstruct the original image based on its features through a reconstruction loss that maximizes the Structural Similarity Index Measure (SSIM) \citep{wang2004ssim} and the Peak Signal-to-Ratio Noise (PSNR). Furthermore, we use an adversarial loss typical of GANs to enhance the quality of the images. To deal with small amounts of data, we follow an approach similar to the work of Karras \textit{et al.} \citep{karras2020training} by implementing augmentations such as random crops, brightness adjustments, and random flips on the generated images before providing them to the discriminator. The discriminator is trained in an adversarial manner to distinguish between real and generated images. The realism loss term is shown in Equation~\ref{eq:realism-loss}, where $I$ represents the original image, and $\alpha$ is a threshold used to normalize the PSNR term of the loss function.

\begin{eqnarray}
\begin{aligned}
\mathcal{L}_{realism} = E_{I \sim p_d(I)}[-\log(C_{real}(D(E(I)))) + (1-SSIM(I, D(E(I)))) + (1-\frac{PSNR(I, D(E(I)))}{\alpha})] \label{eq:realism-loss}
\end{aligned}
\end{eqnarray}

The whole loss function used to train the disentanglement network is exposed in Equation~\ref{eq:whole-loss}, where $\lambda_{r}$ and $\lambda_{d}$ are parameters used to calibrate the relative importance of each loss term.

\begin{eqnarray}
\begin{aligned}
\mathcal{L} = \mathcal{L}_{class} + \lambda_{r} \mathcal{L}_{realism} + \lambda_{d} \mathcal{L}_{disent} \label{eq:whole-loss}
\end{aligned}
\end{eqnarray}

\subsection{Image Anonymization}

To anonymize an image using the proposed disentanglement network, we must alter its vector of identity features. To guarantee that the anonymized images are realistic, the synthetic features that will replace the original images' identity must follow the same probability distribution as real identity features. Furthermore, the synthetic identity features must differ sufficiently from the identities in the training data to avoid identity leaks. Given these requirements, we propose training a VAE to produce synthetic identity features that follow the probability distribution of real data. Furthermore, we implement a privacy loss to protect the privacy of patients that maximizes the entropy of the network on identity features generated from random samples. The loss function to train the VAE is exposed in Equation~\ref{eq:privacy-loss-multiclass}, where $VAE$, $E_{vae}$ and $D_{vae}$ represent the output of the proposed VAE, its encoder and its decoder, respectively, $E_{id}(I)$ represents the identity features of the original image $I$, $KL$ represents the Kullback-Leibler divergence, and $X$ is vector randomly sampled from a normal distribution $N(0,1)$.

\begin{eqnarray}
\begin{aligned}
\mathcal{L}_{privacy} = E_{I \sim p_d(I)}[(E_{id}(I) - VAE(E_{id}(I)))^2) + KL(E_{vae}(E_{id}(I)) \lvert \rvert N(0, 1)) \\+ C_{id}(D_{vae}(X)) \log(C_{id}(D_{vae}(X)))] \label{eq:privacy-loss-multiclass}
\end{aligned}
\end{eqnarray}

In terms of architecture, the encoder and decoder of the proposed VAE are multi-layer perceptrons with one hidden layer and one dropout layer each. 

\begin{figure*}[t]
    \centering
    \includegraphics[scale=0.65]{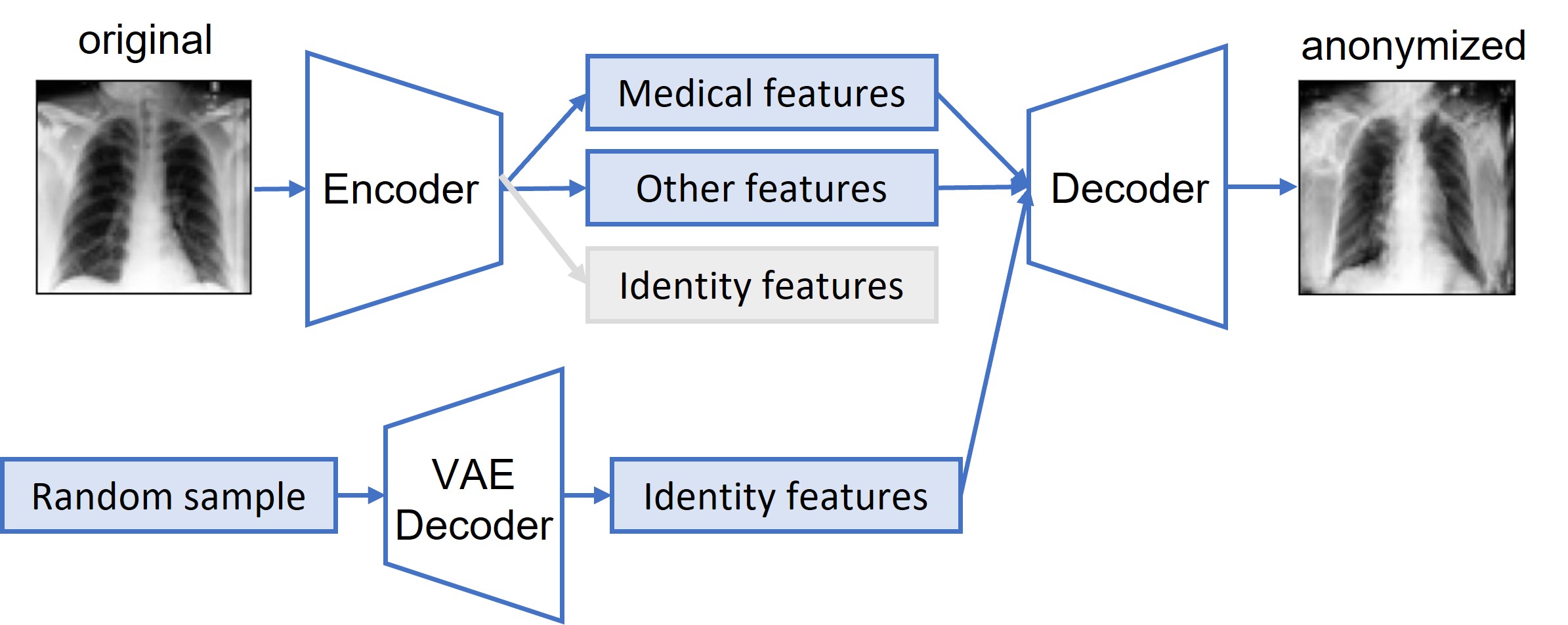}
    \caption{Representation of the anonymization process by sampling a new identity vector using the decoder of a VAE.} \label{fig:anonymization-process}
\end{figure*}

During anonymization, we use the decoder of the VAE to sample a new synthetic identity, which replaces the identity of the original image to produce an anonymized image, as depicted in Figure~\ref{fig:anonymization-process}.

\subsection{Siamese Identity Recognition}

In the previous sections, we defined the identity recognition network that captures identity features as a multi-class classification network. However, when there is a high number of patients in the data and a low number of images per patient, it may be difficult to train a multi-class classification network. In these cases, the multi-class identity recognition network may be replaced by a Siamese network \citep{bromley1993siamese}.

In the disentanglement network, the Siamese network captures identity features by maximizing the distance between identity features from different patients and minimizing this distance between identity features of different images from the same patient. To train a Siamese network, in addition to the two images (original and target) that are given to the disentanglement network during training, we must also provide a third image from the same patient as the original image, exclusively for training the identity recognition network. The classification loss of the disentanglement network in the presence of a Siamese identity recognition network is depicted in Equation~\ref{eq:siamese-classifier}, where $I_{ori}$, $I_{same}$ and $I_{tar}$ represent the original image, an image from the same patient and an image from a different patient, respectively, and $\lambda_{med}$ and $\lambda_{id}$ are parameters to control the importance of each loss term.

\begin{eqnarray}
\begin{aligned}
\mathcal{L}_{class} = E_{I \sim p_d(I)}[-\lambda_{med} y_{med}\log(C_{med}(E_{med}(I_{ori}))) + \lambda_{id} ((E_{id}(I_{ori}) - E_{id}(I_{same}))^2 \\+ \max~(0.1-(E_{id}(I_{ori}) - E_{id}(I_{tar}))^2, 0))] \label{eq:siamese-classifier}
\end{aligned}
\end{eqnarray}

In the anonymization process, instead of maximizing the entropy of a multi-class classification network, the Siamese identity recognition network is used to maximize the distance between the synthetic identity features randomly sampled using the VAE and the identity features of the original image. Furthermore, to protect privacy for all patients in the data, we also maximize the distance between the sampled synthetic features and the identity features of randomly selected patients from the training data. The privacy loss function used to train the VAE using a Siamese identity classifier is depicted in Equation~\ref{eq:privacy-loss-siamese}, where $E_{id}(I)$ and $E_{id}(I_2)$ represent the identity features of the original patient and of a randomly selected patient, respectively.

\begin{eqnarray}
\begin{aligned}
\mathcal{L}_{privacy} = E_{I,I_2 \sim p_d(I)}[(E_{id}(I) - VAE(E_{id}(I)))^2) + KL(E_{vae}(E_{id}(I)) \lvert \rvert N(0, 1)) \\+ \max(0.1-(E_{id}(I) - D_{vae}(X))^2, 0)) + \max(0.1-(E_{id}(I_2) - D_{vae}(X))^2, 0))] \label{eq:privacy-loss-siamese}
\end{aligned}
\end{eqnarray}

\section{Experimental Setup}  \label{sec:experiments}

This section describes the datasets used in the experiments, the models trained for each dataset, and the methods used to evaluate the models.

\subsection{Data}
To evaluate the proposed disentanglement network, we use three datasets:

\begin{itemize}
    \item \textbf{Warsaw-BioBase-Disease-Iris v2.1} \citep{trokielewicz2015iris,trokielewicz2017iris}: contains 1,795 grayscale eye iris images from 115 different patients. The images are annotated for various medical conditions. We focus on glaucoma recognition, for which the dataset is imbalanced since only 24\% of the images contain glaucoma. We split the data into 1,165 images for training, 270 for validation and 360 for testing, maintaining the class ratio in each set.
    \item \textbf{CheXpert} \citep{irvin2019chexpert}: contains 224,316 chest radiographs from  65,240 patients. We use 6,000 images from 1,791 patients for training, 200 from 84 patients for validation, and 300 from 130 patients for testing. We focus on the recognition of Pleural Effusion, for which the selected set of images is balanced.
    \item \textbf{CelebAMask-HQ} \citep{lee2020maskgan}: contains 30,000 high-quality face images from celebrities. We use 8,210 images from 600 celebrities for training, 268 images from 20 celebrities for validation, and 371 images from 30 celebrities for testing. We use the segmentation masks of the dataset to remove the images' background, aiming to decrease the variability in the data and facilitate the proposed model's training. The target task that serves as a proxy for the disease recognition task is facial expression recognition, focusing on recognizing smiling. For this task, the dataset is imbalanced as 57\% of the selected face images are smiling.
\end{itemize}

The biometric nature of iris data facilitates the identity recognition process, which enables the evaluation of the proposed network's ability to anonymize images. However, since these images are not typically used for glaucoma recognition, it is difficult to verify whether medical features are properly preserved during anonymization. Nevertheless, this dataset is used to compare the proposed approach with other works from the literature that use the same dataset.
The medical nature of the chest data enables the evaluation of the network's capacity to disentangle medical features and preserve them through anonymization.
Since the iris and chest datasets are limited in showing the strengths of the disentanglement network due to their difficult-to-observe medical and identity features, respectively, we also use face data to evaluate the network's capacity to anonymize images while retaining relevant explanatory features. Here, we use facial expression recognition as a proxy for the disease recognition task.

\subsection{Training Details}

We used a Siamese identity recognition network in the experiments with the chest and face data, and a multi-class identity recognition network in the experiments with the iris data.

During training, we used the validation set to measure the performance of the models, saving the models with the best performance on validation. The disentanglement network was trained for 1,551 epochs on the chest data, 1,644 epochs on the face data, and 4,635 epochs on the iris data. We used the Adam optimizer with a learning rate of $2 \times 10^{-5}$ in all experiments. The parameters empirically selected for the loss function of the disentanglement network were $\lambda_{med}=5$, $\lambda_{id}=5$, $\lambda_{r}=1$, and $\lambda_{d}=5$ for the chest data, $\lambda_{med}=0.5$, $\lambda_{id}=10$, $\lambda_{r}=0.02$, and $\lambda_{d}=10$ for the face data, and $\lambda_{med}=1$, $\lambda_{id}=1$, $\lambda_{r}=0.1$, and $\lambda_{d}=5$ for the iris data. We used $\alpha=48$ in all experiments. The VAE to generate synthetic identities was trained for 223, 135 and 803 epochs on the chest, face and iris data, respectively.

\subsection{Evaluation}

\begin{figure*}[t]
    \centering
    \includegraphics[scale=0.65]{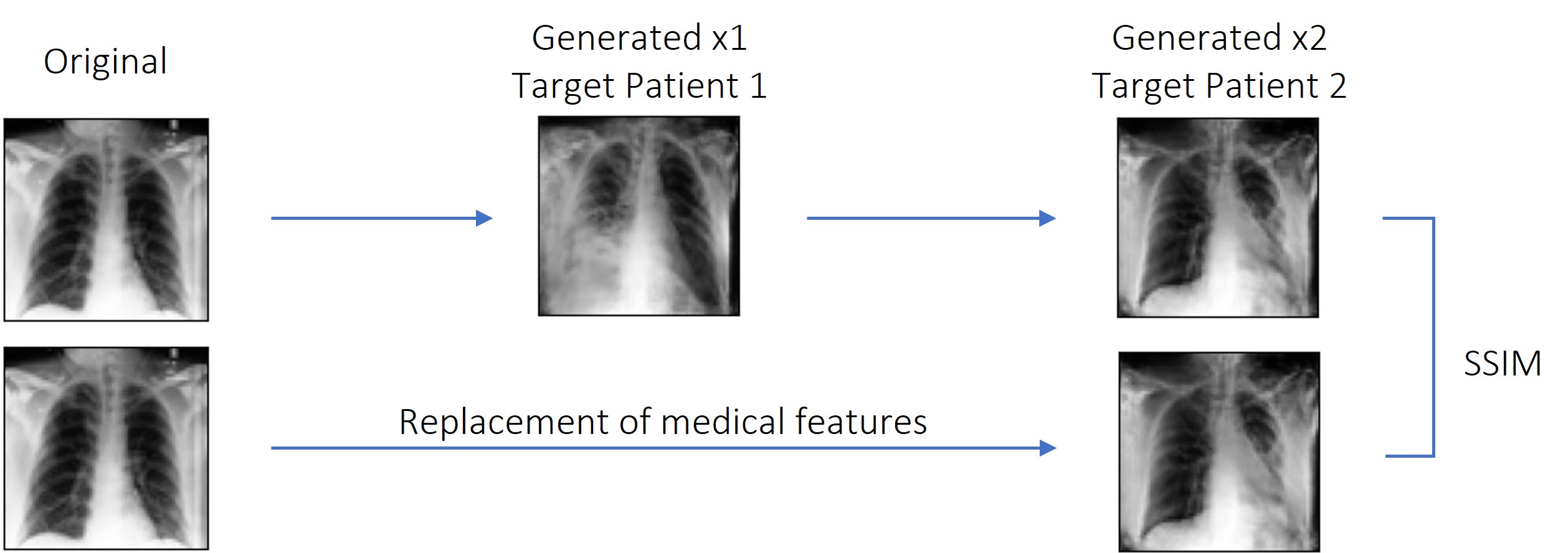}
    \caption{Representation of the evaluation of realism in the process of replacing medical features, by comparing an image whose features were replaced twice with an image whose features were only replaced once by the features of the final target patient.} \label{fig:evaluation-process}
\end{figure*}

The disentanglement network is evaluated in two aspects: realism of the generated images, and capacity to replace medical or identity features while preserving the remaining features. For each test image, we obtain three images that represent its reconstruction, the replacement of its medical features and the replacement of its identity features. Regarding realism, the images are evaluated according to their similarity to the original image, obtained using SSIM. Higher SSIM values during reconstruction indicate better image quality. Furthermore, we compare the generated images with images that went through reconstruction/replacement of features twice, as depicted in Figure~\ref{fig:evaluation-process}, to evaluate how much each iteration of the disentanglement processes affects image quality. 
To evaluate disentanglement, we measure the accuracy of an identity recognition network in recognizing the identities of the original and target patients in the generated images. To compute accuracy using the Siamese identity recognition network, whose loss maximizes and minimizes distances between identity features, we use a threshold ($t = 0.05$) such that two images are considered to belong to the same identity if their distance is less than the threshold. We select the threshold based on the value that was used as the maximum value that the distance should achieve ($0.1$) in the identity recognition loss in the previously shown Equation~\ref{eq:siamese-classifier}. When identity features are replaced, we expect to obtain high values in the recognition of the target identity and low values in the recognition of the original identity. When medical features are replaced, we expect to obtain high values in the recognition of the original identity and low values in the target identity recognition. We also apply the disease recognition network to evaluate whether the class of the generated image matches the class of the original patient when identity is replaced, and of the target patient when medical features are replaced. Thus, we expect high accuracy in disease recognition in all settings.

The anonymization process is also evaluated using the identity and disease recognition networks. To verify whether the privacy of the original patient is preserved in its anonymized image, we observe the accuracy of the identity classifier in attempting to identify the original patient. To measure privacy for all the training data using a Siamese network, we compute an overall accuracy score that compares the anonymized image with images of all patients from the training data. Using a multi-class identity recognition network, we observe the maximum confidence of the network in its prediction. The lower these values are, the more difficult it is for an identity recognition network to identify patients, suggesting that privacy is preserved. We also measure the accuracy of the disease recognition network to verify whether the medical features are preserved during anonymization, expecting to obtain high values.

Finally, we analyze images showing the results of the disentanglement and anonymization processes, verifying whether we can visually recognize the replacement of the medical and biometric features.

\section{Results} \label{sec:results}

The following sections present the results obtained by applying the disentanglement and anonymization networks to the three datasets. Moreover, we present an ablation study that shows the effects of the various terms of the realism loss on the network, using the iris dataset. Finally, we compare the results of using different numbers of images to train the disentanglement network, using the chest dataset.

\subsection{Disentanglement}

\begin{table*}[]
\caption{Disentanglement results. The realism column represents the SSIM score obtained when comparing the images of each experiment with the original image (Original), and with images whose features were replaced in two iterations (Generated x2). Identity recognition measures the accuracy of recognizing the original and target identities in the images. Disease recognition measures Accuracy and F1-Score.}

\label{tab:disentanglement}
\centering
\begin{tabular}{|l|l|c|c|c|c|c|c|}
\hline
\multirow{2}{*}{\textbf{Data}} & \multirow{2}{*}{\textbf{Experiment}} & \multicolumn{2}{c|}{\textbf{Realism SSIM}} & \multicolumn{2}{c|}{\textbf{Identity Accuracy}} & \multicolumn{2}{c|}{\textbf{Disease Recognition}} \\ \cline{3-8} 
& & \textbf{Original} & \textbf{Generated x2} & \textbf{Original} & \textbf{Target} & \textbf{Accuracy} & \textbf{F1-Score} \\ \hline
\multirow{3}{*}{Chest} & Reconstruction / Baseline & 65.46\% & 96.27\% & 77.24\% & 13.95\% & 81.73\% & 82.65\%  \\ \cline{2-8}
 & Medical Replacement & 46.22\% & 97.35\% & 100.00\% & 11.63\% & 73.42\% & 68.75\% \\ \cline{2-8}
 & Identity Replacement & 61.55\% & 96.22\% & 12.29\% & 100.00\% & 76.41\% & 72.37\% \\ \hline
\multirow{3}{*}{Face} & Reconstruction / Baseline & 59.30\% & 90.68\% & 93.26\% & 2.16\% & 93.80\% & 92.60\%  \\ \cline{2-8}
 & Medical Replacement & 46.63\% & 92.17\% & 100.00\% & 1.62\% & 92.72\% & 88.80\% \\ \cline{2-8}
 & Identity Replacement & 55.52\% & 90.37\% & 2.16\% & 100.00\% & 92.99\% & 91.56\% \\ \hline
\multirow{3}{*}{Iris} & Reconstruction / Baseline & 53.36\% & 98.36\% & 81.76\% & 0.00\% & 90.00\% & 81.91\%  \\ \cline{2-8}
 & Medical Replacement & 52.96\% & 98.40\% & 82.35\% & 0.59\% & 74.41\% & 78.83\% \\ \cline{2-8}
 & Identity Replacement & 29.01\% & 99.86\% & 0.00\% & 82.65\% & 85.88\% & 71.76\% \\ \hline
\end{tabular}
\end{table*}

Table~\ref{tab:disentanglement} presents the results of the disentanglement network on the three datasets, in three different settings: reconstruction, replacement of medical features and replacement of identity features. It shows the SSIM between the images generated in each experiment and the original image, as well as with images that went through the experiment twice (Generated x2). It also shows the results of the identity recognition network applied to identify the identities of the original and target patients in the images. Finally, it shows the results of the disease recognition network applied to identify the class that the generated image should have. The experiment referring to reconstruction shows the results of realism applied to the reconstructed images and the results of the identity and disease recognition networks applied to the original images (baseline).

We find that the network achieves higher SSIM between the original and generated images in the chest data. Nevertheless, for the iris data, applying the disentanglement model twice leads to a minimal loss of image quality. Considering the identity replacement results, the identity recognition network can recognize the target identity with high accuracy and obtains low accuracy at recognizing the original patient, indicating that the identity replacement process is successful. Furthermore, the disease recognition networks achieve high accuracy for all datasets in this setting, suggesting that medical features are preserved through the replacement of identity features. In medical feature replacement, the disease recognition network obtains high accuracy in identifying the target medical condition. Nevertheless, it seems to be particularly difficult to alter the medical features in the iris dataset, as evidenced by the higher difference between the accuracy results in the baseline and in medical replacement. The identity classifier is capable of identifying the original patient and fails at identifying the target patient, suggesting that the alteration of medical features does not affect identity.
Overall, the face dataset seems to be the easiest to disentangle, perhaps due to its well-defined identity and facial recognition features.

\begin{figure*}[t!]
    \centering
    \includegraphics[scale=0.7]{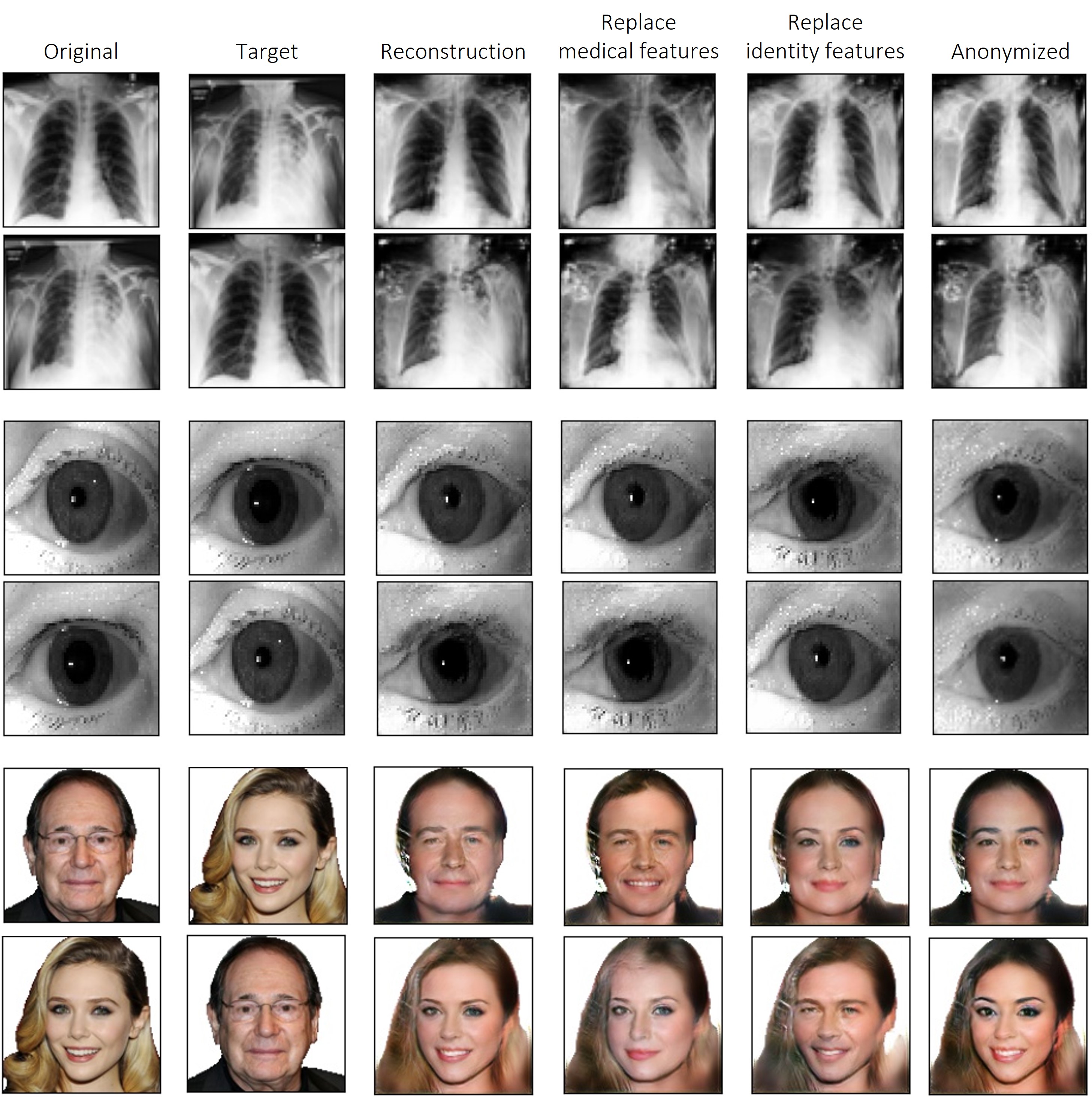}
    \caption{Results of disentanglement and anonymization on the chest data, iris data and face data.} \label{fig:disentanglement}
\end{figure*}

Figure~\ref{fig:disentanglement} displays visual results of disentanglement. In the iris dataset, the reconstructed images and the images whose medical features are replaced seem very similar, making it difficult to verify whether the disentanglement of medical features was successful. Nevertheless, by looking at the chest and face data, we can observe that the network is capable of replacing the medical and identity features of the original images with features that are visually similar to those of the target image. In the chest data, alterations to the identity of the network alter the shape of the torso without altering its position. The medical and identity features seem to be independent of the patient's position. In a similar manner, the position of the celebrities in the face dataset is preserved through alterations to identity and facial expression features. In the face data, the network has difficulty reconstructing features like glasses due to the small number of training images that contain glasses (305 images).

The disentanglement network is inherently capable of generating counterfactual images through the replacement of the medical features of the original patient, showing how a patient would look with a different pathology.

\subsection{Image Anonymization}

\begin{table*}[t]
\caption{Anonymization Results. Identity recognition measures the accuracy of a network at recognizing the original and any identity from the dataset in the anonymized images. Disease recognition measures the preservation of medical features through the accuracy of a classifier. The results of the proposed model are highlighted in bold.}
\label{tab:anonymization}
\centering
\begin{tabular}{|l|l|cc|c|}
\hline
\multirow{2}{*}{\textbf{Data}} & \multirow{2}{*}{\textbf{Method}} & \multicolumn{2}{c|}{\textbf{Identity Recognition}} & \textbf{Disease Rec.} \\ \cline{3-4} & & \multicolumn{1}{c|}{\textbf{Original Accuracy}} & \textbf{Overall Accuracy} & \textbf{Accuracy}  \\ \hline
\multirow{2}{*}{Chest} & VAE with overall privacy & \multicolumn{1}{c|}{\textbf{10.63\%}} & \textbf{11.39\%}  & \textbf{78.41\%} \\ \cline{2-5}
  & Model of \cite{montenegro2023disentangled} & \multicolumn{1}{c|}{29.90\%} & 13.83\%  & 74.42\% \\ \hline
Face & VAE with overall privacy & \multicolumn{1}{c|}{\textbf{14.82\%}} & \textbf{12.96\%}  & \textbf{91.64\%} \\ \hline
\multirow{7}{*}{Iris} & VAE with overall privacy & \multicolumn{1}{c|}{\textbf{0.59\%}} & \textbf{11.76\%}  & \textbf{90.29\%} \\ \cline{2-5}
& VAE (without maximum entropy) & \multicolumn{1}{c|}{1.76\%} & 38.99\%  & 88.82\%\\ \cline{2-5}
& Averaging 6 identities & \multicolumn{1}{c|}{0.88\%} & 35.78\%  & 91.76\% \\ \cline{2-5}
& Averaging 12 identities & \multicolumn{1}{c|}{0.59\%} & 21.98\%  & 90.29\% \\ \cline{2-5}
& Averaging all identities & \multicolumn{1}{c|}{2.06\%} & 6.76\%  & 91.18\% \\ \cline{2-5}
& PP-SIR model of \cite{montenegro2021privacy} & \multicolumn{1}{c|}{4.41\%} & 64.07\%  & 79.71\% \\ \cline{2-5}
& Model of \cite{montenegro2023disentangled} & \multicolumn{1}{c|}{0.29\%} & 65.14\%  & 61.47\% \\ \hline

\end{tabular}
\end{table*}

Table~\ref{tab:anonymization} presents the results of the proposed anonymization method on the three datasets, highlighted in bold. Furthermore, the table shows the results of works from the literature on the chest and iris data. Using the iris data, we also compare the proposed VAE with a typical VAE that does not maximize the entropy of an identity recognition network on the sampled identity features, and with a common technique from the literature to anonymize through averaging identities from the training data. 
The visual results of the proposed anonymization network were provided along with the results of disentanglement in Figure~\ref{fig:disentanglement}. 

The proposed VAE is capable of generating synthetic identities where the identity recognition network has difficulty identifying the original identity and other identities from the training data for all the datasets, as evidenced by the low accuracy values in identity recognition. In the chest data, the proposed method surpasses the work from the literature, achieving lower accuracy at recognizing the original patient and higher accuracy at classifying the patient's medical condition.

\begin{figure}[H]
    \centering
    \includegraphics[scale=0.5]{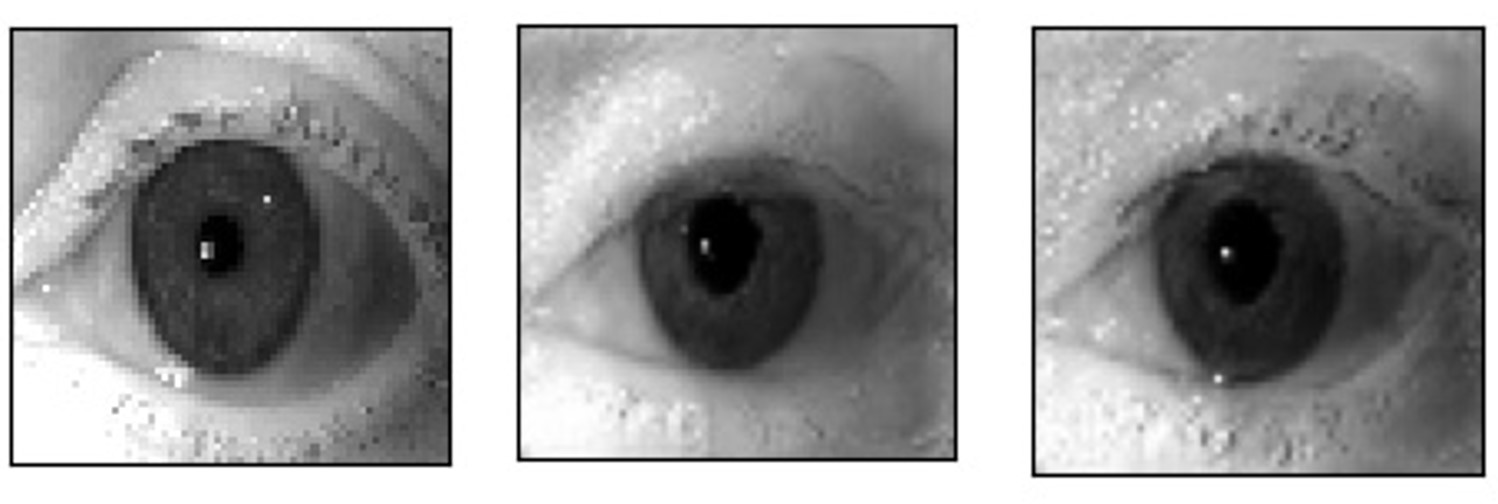}
    \caption{Results of anonymization using the proposed VAE on the iris data. From left to right, the images correspond to the original image, and images obtained with the VAE trained without and with entropy maximization.} \label{fig:anon-iris}
\end{figure}

The experiments with the iris data show the disentanglement network's capacity to preserve the medical condition of the patients in the anonymized images, achieving a high disease recognition accuracy. Maximizing the entropy of the identity recognition network in the loss function of the VAE that synthesizes new identities leads to better overall privacy than using a typical VAE without entropy maximization, as the confidence of the identity recognition network in its prediction is lower in the proposed VAE. Figure~\ref{fig:anon-iris} compares anonymized images obtained with the proposed VAE and a typical VAE.

\begin{figure}[H]
    \centering
    \includegraphics[scale=0.5]{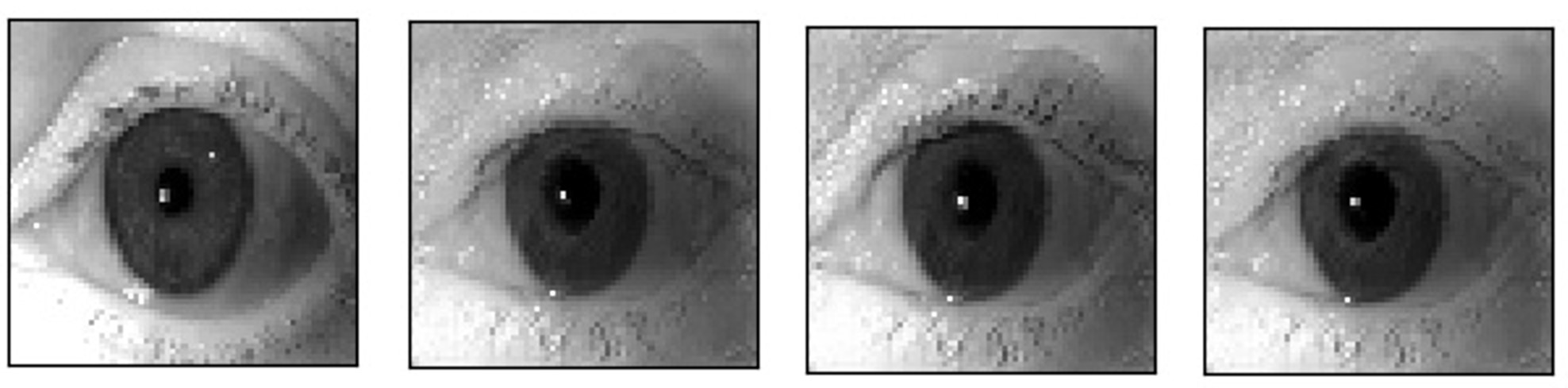}
    \caption{Results of averaging identities as an anonymization technique. From left to right, the images correspond to the original image, and to images anonymized using 6, 12 and all identities, respectively.} \label{fig:averaging-iris}
\end{figure}

Anonymizing through averaging identities from the training data potentially compromises the identities of the patients which are directly used in the anonymization process. Nonetheless, we verify that, as we increase the number of identities used, the more difficult it gets for the identity classifier to identify a patient from the data, as its confidence decreases. Averaging all the identities in the dataset provided competitive results as the proposed VAE identity generator, even surpassing it in the overall privacy metric. Figure~\ref{fig:averaging-iris} shows the visual results produced using this anonymization technique.

We also observe that the proposed network surpasses the literature in regards to both privacy and preservation of medical content in the iris data.

\subsection{Ablation Study}

\begin{figure}[h]
    \centering
    \includegraphics[scale=0.5]{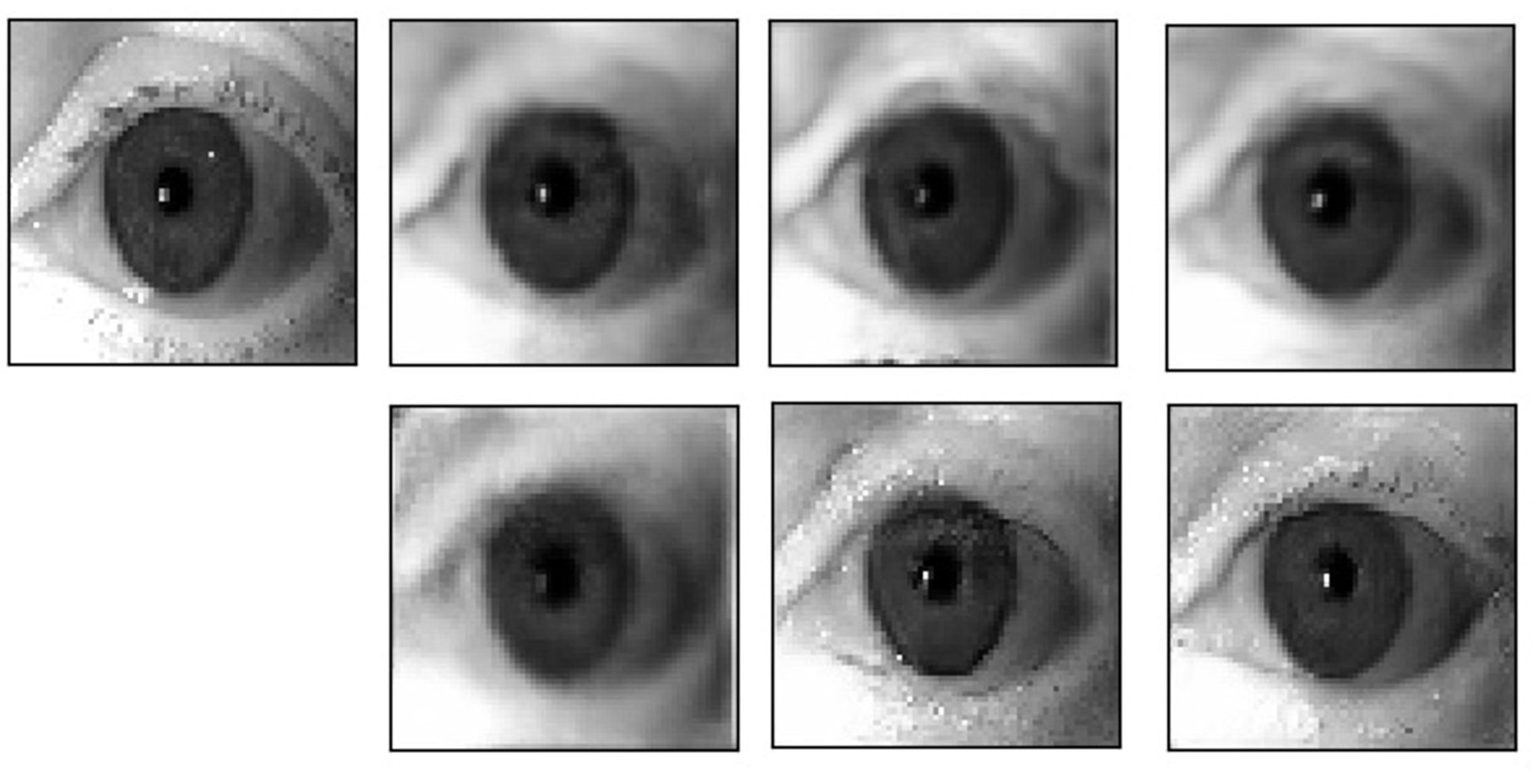}
    \caption{Ablation Study. From left to right, the images correspond to the original and to generated using the network trained only with SSIM, only with PSNR, SSIM and PSNR, adversarial loss in addition to SSIM and PSNR, the fine-tuned and the final model.} \label{fig:ablation-iris}
\end{figure}

We perform an ablation study where we verify the effects of each realism loss term on the realism of the image. We perform various experiments where we train the disentanglement network while making alterations to its realism loss terms. We train the network without the discriminator, using only SSIM, only PSNR, and both SSIM and PSNR. Then, we add a discriminator and an adversarial loss, training the network as a typical GAN without the augmentations on the generated images. Then, we use the model trained using only SSIM and PSNR and fine-tune it by adding the adversarial loss and augmentations on the generated images. Finally, we compare all these methods with the final network, which was trained from scratch using the realism loss with all its terms as depicted in Equation~\ref{eq:realism-loss}. The results of the ablation study are presented in Table~\ref{tab:ablation-iris}, which evaluates the SSIM between the original and reconstructed images obtained using each model. Figure~\ref{fig:ablation-iris} displays reconstructed images obtained with each model represented in the ablation study.

\begin{table}[H]
\caption{Ablation Study comparing the realism of the images by altering the realism loss in the model's training.}
\label{tab:ablation-iris}
\centering
\begin{tabular}{|l|c|}
\hline
\textbf{Realism Loss} & \textbf{SSIM} \\ \hline
Only SSIM & 68.26\% \\ \hline
Only PSNR & 66.91\% \\ \hline
SSIM and PSNR & 70.01\% \\ \hline
GAN with SSIM and PSNR  & 57.43\% \\ \hline
Fine-tuned augmented GAN & 60.73\% \\ \hline
Augmented GAN (Final model) & 53.36\% \\ \hline
\end{tabular}
\end{table}

Despite the networks that do not use the adversarial loss of a GAN obtaining higher SSIM, we can observe in the visual results that the images generated through these methods are blurry. Adding a GAN loss without augmentations on the generated images also leads to blurry images and a significant decrease in structural similarity, perhaps due to the adversarial loss of the GAN leading the generator to add features that do not exist in the original image in its attempt to enhance the realism of the images and bring the probability distribution of the generated images closer to that of real images. The methods where we add the augmentations to the generated images in the network lead to higher-quality results. In the visual results, the fine-tuned network seems to lead to slightly lower-quality images with more artefacts when compared with the final network trained from scratch, despite achieving higher structural similarity between original and reconstructed images.

\begin{figure}[h]
    \centering
    \includegraphics[scale=0.6]{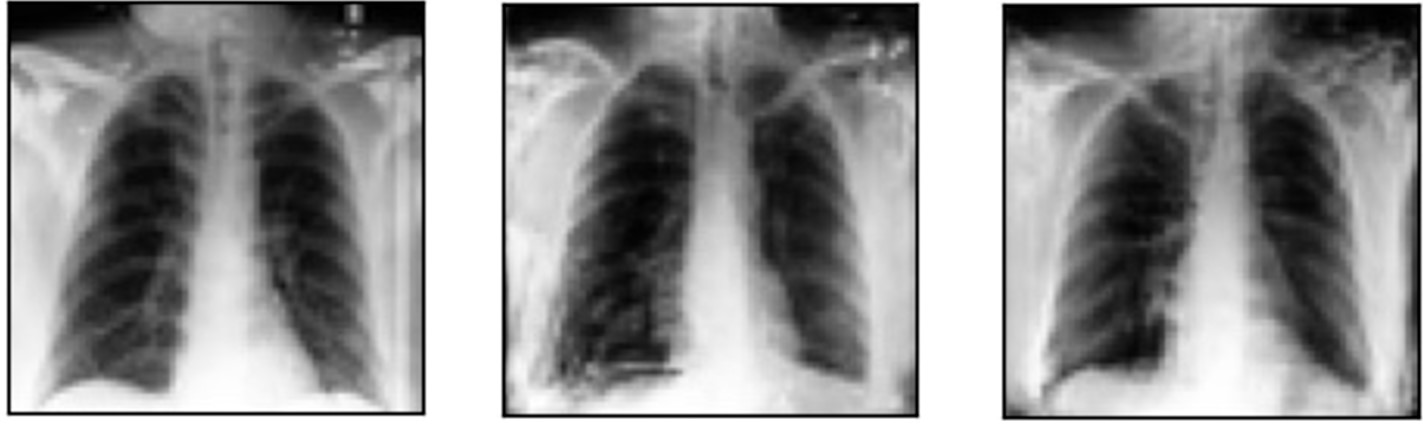}
    \caption{Ablation Study showing (from left to right) the original image, and images obtained with the network trained with 3,000, and 6,000 images.} \label{fig:ablation-chexpert}
\end{figure}

We also perform an ablation study that analyzes the effects of training the disentanglement model with different amounts of data. Table~\ref{tab:ablation-chexpert} and Fig~\ref{fig:ablation-chexpert} display these results.

\begin{table}[h]
\caption{Ablation Study comparing results when differently sized datasets are used to train the network. ``N.~Images'' represents the number of images used during training, and ``Rec.'' stands for Recognition.}
\label{tab:ablation-chexpert}
\centering
\begin{tabular}{|c|c|c|c|}
\hline
\textbf{N. Images} & \textbf{SSIM} & \textbf{Identity Rec.} & \textbf{Disease Rec.} \\ \hline
3,000 & 65.64\% & 65.78\% & 77.08\% \\ \hline
6,000 & 65.46\% & 77.24\% & 81.73\% \\ \hline
\end{tabular}
\end{table}

We observe that it is possible to obtain comparable levels of realism in the models trained with 3,000 and 6,000 images. However, the identity and pleural effusion recognition networks achieve lower accuracy when trained with less data.

\section{Conclusions}  \label{sec:conclusions}

This work proposes a novel methodology to disentangle identity and medical features from medical images inspired by the definition of independence between variables. The disentanglement model was applied to the anonymization of medical and biometric images by replacing the images' identity features. We propose a variational autoencoder that generates identity features that protect the privacy of the patients in the training data.
The experiments demonstrate that the proposed network successfully disentangles identity and medical characteristics in images. Furthermore, the experiments show that the proposed network surpasses the state-of-the-art anonymization methods for medical case-based explanations.

The experiments show the proposed model's inherent capacity to produce counterfactual images through the replacement of medical features. Future work will focus on disentangling the medical features into specific variables causally related to the disease prediction task, to enable a more meaningful and controlled generation of counterfactual explanations.

To conclude, anonymizing medical case-based explanations is imperative to incorporate deep learning-based decision support systems that use case-based reasoning, including medical image retrieval systems, into clinical practice. This work contributes towards increasing the safety of patients and the trust of medical experts in using such models, enabling their deployment into the real world.

\section*{Acknowledgments}
This work is financed by National Funds through the FCT - Fundação para a Ciência e a Tecnologia, I.P. (Portuguese Foundation for Science and Technology) within the project CAGING, with reference 2022.10486.PTDC, and within PhD grant with reference 2022.14516.BD.

\bibliographystyle{unsrtnat}
\bibliography{refs}  

\end{document}